\documentclass[sigconf]{acmart}

\AtBeginDocument{%
	\providecommand\BibTeX{{%
			\normalfont B\kern-0.5em{\scshape i\kern-0.25em b}\kern-0.8em\TeX}}}
		
\copyrightyear{2020}
\acmYear{2020}
\setcopyright{acmcopyright}\acmConference[MM '20]{Proceedings of the 28th ACM International Conference on Multimedia}{October 12--16, 2020}{Seattle, WA, USA}
\acmBooktitle{Proceedings of the 28th ACM International Conference on Multimedia (MM '20), October 12--16, 2020, Seattle, WA, USA}
\acmPrice{15.00}
\acmDOI{10.1145/3394171.3413973}
\acmISBN{978-1-4503-7988-5/20/10}

\usepackage{balance}
\usepackage{booktabs} 
\usepackage{graphicx}
\usepackage{subfigure}
\usepackage{bm}
\usepackage{multirow}
\usepackage{bbding} 
\usepackage{algorithm}
\usepackage{algorithmic}

\begin{document}
	\fancyhead{}
	
	\title{Cloze Test Helps: Effective Video Anomaly Detection via Learning to Complete Video Events}

\author{Guang Yu}
\authornote{Authors contributed equally to this work.}
\author{Siqi Wang}
\authornotemark[1]
\affiliation{%
	\institution{National University of Defense Technology}
}
\email{{yuguangnudt, wangsiqi10c}@gmail.com}

\author{Zhiping Cai}
\affiliation{%
	\institution{National University of Defense Technology}
}
\email{zpcai@nudt.edu.cn}

\author{En Zhu}
\affiliation{%
	\institution{National University of Defense Technology}
}
\email{enzhu@nudt.edu.cn}

\author{Chuanfu Xu}
\affiliation{%
	\institution{National University of Defense Technology}
}
\email{xuchuanfu@nudt.edu.cn}

\author{Jianping Yin}
\affiliation{%
	\institution{Dongguan University of Technology}
}
\email{jpyin@dgut.edu.cn}

\author{Marius Kloft}
\affiliation{%
	\institution{TU Kaiserslautern}
}
\email{kloft@cs.uni-kl.de}

\renewcommand{\shortauthors}{Yu and Wang, et al.}

\begin{abstract}
	As a vital topic in media content interpretation, video anomaly detection (VAD) has made fruitful progress via deep neural network (DNN). However, existing methods usually follow a reconstruction or frame prediction routine. They suffer from two gaps: (1) They cannot localize video activities in a both precise and comprehensive manner. (2) They lack sufficient abilities to utilize high-level semantics and temporal context information. Inspired by frequently-used \textit{cloze test} in language study, we propose a brand-new VAD solution named \textit{Video Event Completion} (VEC) to bridge gaps above: First, we propose a novel pipeline to achieve both precise and comprehensive enclosure of video activities. Appearance and motion are exploited as mutually complimentary cues to localize regions of interest (RoIs). A normalized spatio-temporal cube (STC) is built from each RoI as a \textit{video event}, which lays the foundation of VEC and serves as a basic processing unit. Second, we encourage DNN to capture high-level semantics by solving a \textit{visual cloze test}. To build such a visual cloze test, a certain patch of STC is erased to yield an incomplete event (IE). The DNN learns to restore the original video event from the IE by inferring the missing patch. Third, to incorporate richer motion dynamics, another DNN is trained to infer erased patches' optical flow. Finally, two ensemble strategies using different types of IE and modalities are proposed to boost VAD performance, so as to fully exploit the temporal context and modality information for VAD. VEC can consistently outperform state-of-the-art methods by a notable margin (typically $1.5\%$--$5\%$ AUROC) on commonly-used VAD benchmarks. Our codes and results can be verified at \textit{\url{github.com/yuguangnudt/VEC_VAD}}.
\end{abstract}

\ccsdesc[500]{Computing methodologies~Scene anomaly detection}
\ccsdesc[500]{Computing methodologies~Unsupervised learning}

\keywords{Video anomaly detection, video event completion}

\maketitle

\section{Introduction}
\label{sec:intro}

\begin{figure*}
	\centering
	\includegraphics[scale=1.18]{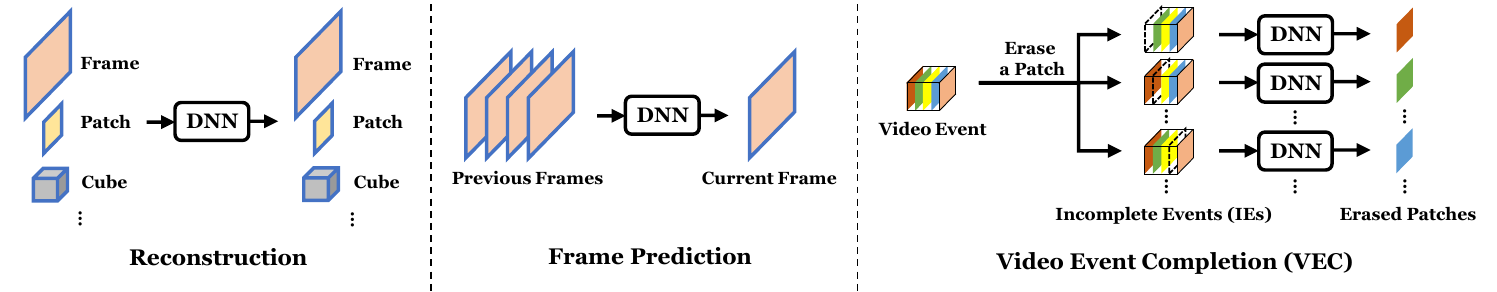}
	\caption{Typical solutions for DNN based VAD. \textbf{\textit{Left}}: Reconstruction based methods train DNN to reconstruct inputs, e.g. video frames/patches/cubes. \textbf{\textit{Middle}}: Frame prediction based methods take several previous frames as inputs of DNN to predict current frame. \textbf{\textit{Right}}: VEC first encloses video events with spatio-temporal cubes (STCs) based on a novel pipeline that synthesizes appearance and motion cues. Then, erasing the patch at STC's different positions produces different types of incomplete events (IEs), which serves as different ``visual cloze tests''. Each DNN is trained to solve a visual cloze test, i.e. learning to complete the missing patch of a certain type of IE. Note that cubes for reconstruction differs from STCs in VEC, as they are yielded by a relatively coarse strategy (e.g. sliding windows) and cannot enclose video events both precisely and comprehensively.}
	\label{fig:recon_pre_vec}
\end{figure*}

Videos play a key role in multimedia. Video anomaly detection (VAD), which performs anomaly detection by interpreting the video content automatically, has been appealing to both academia and industry, since it is valuable to various safety-critical scenarios like municipal and traffic management. Formally, VAD refers to detecting video activities that divert significantly from the observed normal routine. Despite many efforts, VAD remains challenging for two features of anomaly \cite{chandola2009anomaly}: \textbf{(1)} \textit{Scarcity}. As anomalies are usually rare, collecting real anomalies for training is often hard or even impossible. \textbf{(2)} \textit{Ambiguity}. The anomaly does not possess fixed semantics and may refer to different activities based on different context, so it can be highly variable and unpredictable. Such features render modeling anomalies directly unrealistic. Thus, VAD usually adopts an \textit{one-class classification} setup \cite{khan2014one}. This setup collects training videos with only normal activities, which are much more accessible than anomalies, to build a normality model. Activities that do not conform to this model are then viewed as anomalies. As all training data are normal, discriminative supervised learning is usually not applicable. Instead, the unsupervised/self-supervised learning has been the commonly-used scheme in VAD. Following such a scheme, existing VAD solutions fall into two categories: \textbf{(1)} \textit{Classic} VAD, which requires domain knowledge to design hand-crafted descriptors to depict high-level features (e.g. trajectory, speed) or low-level features (e.g. gradient, texture) of video activities. Extracted features are fed into classic anomaly detection methods like one-class support vector machine (OCSVM) to spot anomalies. Feature engineering of classic VAD can be labor-intensive and sub-optimal \cite{xu2017detecting}, and designed descriptors are often hard to transfer among different scenes. \textbf{(2)} \textit{Deep neural network (DNN)} based VAD, which is inspired by DNN's success in traditional vision tasks \cite{lecun2015deep}. Due to DNNs' strong capabilities in feature learning and activity modeling \cite{hasan2016learning}, DNN based VAD achieves superior performance and enjoys surging popularity when compared with classic VAD.

Despite the fruitful progress, DNN based VAD still suffers from two gaps. \textbf{Gap \#1:} Existing DNN based VAD methods cannot localize video activities in a both precise and comprehensive manner. A standard practice in VAD is to use a sliding window with motion filtering \cite{xu2017detecting,sabokrou2018deep}, but such localization is obviously imprecise. Recent DNN based VAD methods like \cite{liu2018future,nguyen2019anomaly,ye2019anopcn} simply ignore this issue by learning on the whole frame, but this suffers from scale variations incurred by image depth and foreground-background imbalance problem \cite{liu2019exploring,zhou2019attention}. Few studies \cite{hinami2017joint,ionescu2019object} improve localization precision by a pre-trained object detector, but it causes a ``\textit{closed world}'' problem--The detector only recognize objects in training data and tends to omit video novelties, which leads to non-comprehensive localization results. Such a gap degrades later video activitiy modeling; \textbf{Gap \#2:} Existing DNN based methods lack sufficient abilities to exploit high-level semantics and temporal context in video activities. As illustrated by Fig. \ref{fig:recon_pre_vec}, two paradigms (\textit{reconstruction} and \textit{frame prediction}) dominate DNN based VAD in the literature: Reconstruction based methods learn to reconstruct inputs and detect poorly reconstructed data as anomalies. However, in this case DNNs tend to memorize low-level details rather than learning high-level semantics \cite{larsen2016autoencoding}, and they even reconstruct anomalies well due to overly strong modeling power \cite{Gong_2019_ICCV}; Frame prediction based methods learn to predict a normal video frame from previous video frames, and detect poorly predicted frames as anomalies. Prediction makes it hard to simply memorize details for reducing training loss, but it scores anomalies by the prediction error of a single frame, which overlooks the temporal context. Thus, neither reconstruction nor frame prediction provides a perfect solution. Unlike recent research that focuses on exploring better network architectures to improve reconstruction or frame prediction, we are inspired by \textit{cloze test} in language study and mitigate gaps above by proposing \textit{Video Event Completion} (VEC) as a new DNN based VAD solution (see Fig. \ref{fig:recon_pre_vec}). Our contributions are summarized below:

\begin{itemize}
	\item VEC for the first time combines both appearance and motion cues to localize video activities and extract video events. It overcomes the ``closed world'' problem and enables both precise and comprehensive video activity enclosure, and it lays a firm foundation for video event modeling in VEC.
	
	\item VEC for the first time designs visual cloze tests as a new learning paradigm, which trains DNNs to complete the erased patches of incomplete video events, to substitute frequently-used reconstruction or frame prediction based methods.
	
	\item VEC also learns to complete the erased patches' optical flow, so as to integrate richer information of motion dynamics.
	
	\item VEC utilizes two ensemble strategies to fuse detection results yielded by different types of incomplete events and data modalities, which can further boost VAD performance.
\end{itemize}

\section{Related Work}
\label{sec:related_work}
\quad\textbf{Classic VAD.}
Classic VAD usually consists of two stages: Feature extraction by hand-crafted descriptors and anomaly detection by classic machine learning methods. As to feature extraction, early VAD methods usually adopt tracking \cite{lan2020semi} to extract high-level features like motion trajectory \cite{zhang2009learning, piciarelli2008trajectory} and destination \cite{basharat2008learning}. However, they are hardly applicable to crowded scenes \cite{mahadevan2010anomaly}. To this end, low-level features are extensively studied for VAD, such as dynamic texture \cite{mahadevan2010anomaly}, histogram of optical flow \cite{cong2011sparse}, spatio-temporal gradients \cite{lu2013abnormal, kratz2009anomaly}, 3D SIFT \cite{cheng2015video}, etc. Afterwards, various classic machine learning methods are explored to perform anomaly detection, such as probabilistic models \cite{mahadevan2010anomaly,cheng2015video,antic2011video}, sparse coding and its variants \cite{cong2011sparse,lu2013abnormal,zhao2011online}, one-class classifier \cite{yin2008sensor}, sociology inspired models \cite{Mehran2009Abnormal}. However, feature extraction is the major bottleneck for classic VAD: Manual feature engineering is complicated and labor-intensive, while the designed descriptors often suffer from limited discriminative power and poor transferability among different scenes.

\textbf{DNN Based VAD.}
DNN based VAD differs from classic VAD by learning features automatically from raw inputs with DNNs. The learned features are fed into a classic model or embedded into DNNs for end-to-end VAD. With only normal videos for training, existing DNN based VAD basically falls into a reconstruction or frame prediction routine. They are reviewed respectively below: \textbf{(1)} \textit{Reconstruction} based methods learn to reconstruct inputs from normal training videos, and assume that a large reconstruction error signifies the anomaly. Autoencoder (AE) and its variants are the most popular DNNs to perform reconstruction. For example, \cite{xu2015learning} pioneers DNN based VAD by introducing stacked denoising AE (SDAE) and propose its improvement \cite{xu2017detecting}; \cite{hasan2016learning} adopts convolutional AE (CAE) that are more suitable for modeling videos, while recent works explore numerous CAE variants such as Winner-take-all CAE (WTA-CAE) \cite{tran2017anomaly} and Long Short Term Memory based CAE (ConvLSTM-AE) \cite{luo2017remembering}; \cite{yan2018abnormal} integrates variational AE into a two-stream recurrent framework (R-VAE) to realize VAD; \cite{abati2019latent} equips AE with a parametric density estimator (PDE-AE) for anomaly detection; \cite{Gong_2019_ICCV} propose a memory-augmented AE (Mem-AE) to make AE's reconstruction error more discriminative. In addition to AE, other types of DNNs are also used for the reconstruction purpose, e.g. sparse coding based recurrent neural network (SRNN) \cite{luo2017revisit} and generative adversarial network (GAN) \cite{ravanbakhsh2017abnormal,sabokrou2018adversarially}. Cross-modality reconstruction is also shown to produce good VAD performance by learning the appearance-motion correspondence (AM-CORR) \cite{nguyen2019anomaly}. \textbf{(2)} \textit{Frame prediction} based methods learn to predict current frames by several previous frames, while a poor prediction is viewed as abnormal. \cite{liu2018future} for the first time formulates frame prediction as an independent VAD method, and imposes appearance and motion constraints for prediction quality. \cite{lu2019future} improves frame prediction by using a convolutional variational RNN (Conv-VRNN). However, prediction by a per-frame basis leads to a bias to background \cite{liu2019exploring}, and \cite{zhou2019attention} proposes attention mechanism to ease the issue. Another natural idea is to combine prediction with reconstruction as a hybrid VAD solution: \cite{zhao2017spatio} design a spatio-temporal CAE (ST-CAE), in which an encoder is followed by two decoders for reconstruction and prediction purpose respectively; \cite{morais2019learning} reconstructs and predicts human skeletons by a message-passing encoder-decoder RNN (MPED-RNN); \cite{ye2019anopcn} integrates reconstruction into prediction by a predictive coding network based framework (AnoPCN); \cite{tang2020integrating} conducts prediction and reconstruction in a sequential manner.

\section{Video Event Completion (VEC)}
\label{sec:method}
\subsection{Video Event Extraction}

In this paper, video event extraction aims to enclose video activities by numerous normalized spatio-temporal cubes (STCs). A STC is then viewed as a video event, which serves as the basic processing unit in VEC. A both \textbf{\textit{precise}} and \textbf{\textit{comprehensive}} localization of video activities is the key to video event extraction. Ideally, \textit{precise} localization expects the subject of a video activity is intactly extracted with minimal irrelevant background, while \textit{comprehensive} localization requires all subjects associated with video activities are extracted. As explained by \textbf{Gap \#1} (Sec. \ref{sec:intro}) and the intuitive comparison in Fig. \ref{fig:visual_vee}, existing DNN based VAD methods fail to realize precise and comprehensive localization simultaneously, which undermines the quality of video activity modeling and VAD performance. Hence, we exploit appearance and motion as mutually complementary cues to achieve both precise and comprehensive localization (Fig. \ref{fig:visual_vee} (d)). This new pipeline is detailed as follows:

\begin{figure}
	\centering
	\includegraphics[scale=0.82]{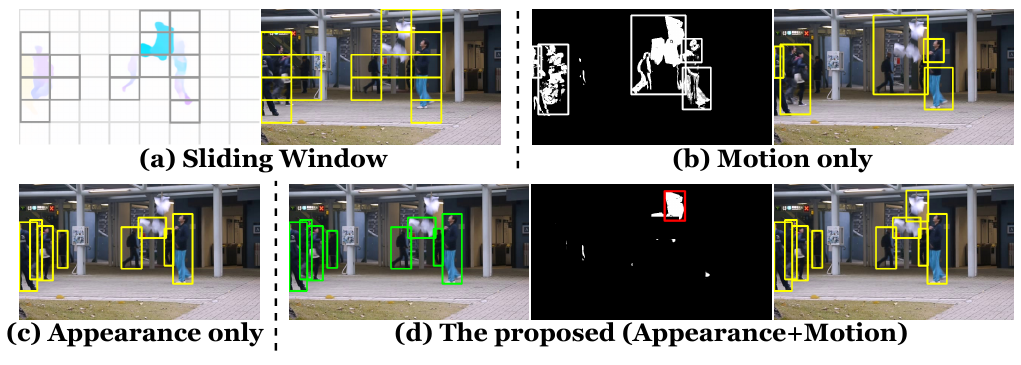}
	\caption{Comparison of localization strategies: Sliding window (a) or motion only (b) produces imprecise localization, while appearance only (c) yields non-comprehensive localization results. The proposed (d) pipeline achieves more precise and comprehensive localization simultaneously.}
	\label{fig:visual_vee}
\end{figure}

\begin{figure*}
	\centering
	\includegraphics[scale=1]{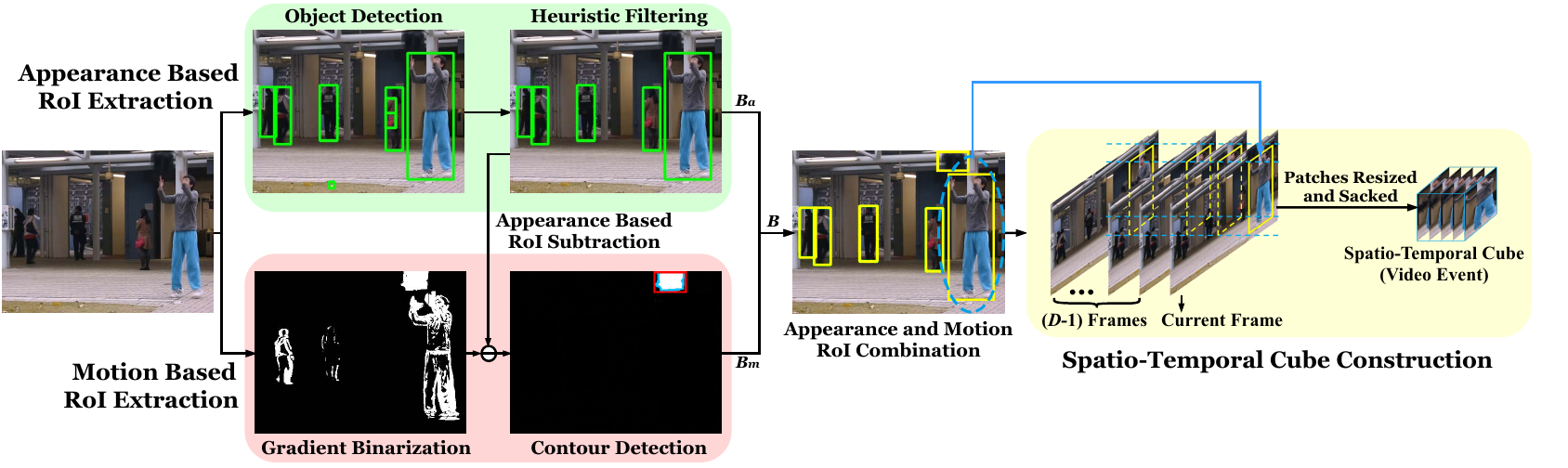}
	\caption{Pipeline of video event extraction: \textbf{(1)} Appearance based RoI extraction (green): Appearance based RoIs are extracted with a pre-trained object detector and filtered based on efficient heuristic rules. \textbf{(2)} Motion based RoI extraction (red): First, temporal gradients are binarized by magnitude into a binary map. Then, highlighted pixels in appearance based RoIs are subtracted from the binary map. Finally, contour detection and simple heuristics are applied to the binary map for final motion based RoIs. \textbf{(3)} Spatio-temporal cube (STC) extraction (yellow): For each RoI, corresponding patches from current frame and $(D-1)$ previous frames are extracted. $D$ patches are then resized and stacked into a STC, which represents a video event.}
	\label{fig:event_extraction}
\end{figure*}

\textbf{Motivation.} As video activities are behaviors conducted by certain subjects in videos, we consider both \textit{appearance cues} from those subjects and \textit{motion cues} from their behaviors to localize regions of interest (RoIs) that enclose those video activities. To utilize appearance cues, a natural solution is modern object detection \cite{cai2018cascade}. With pre-training on large-scale public datasets like Microsoft COCO, pre-trained detectors can precisely localize RoIs with frequently-seen objects. Therefore, we use a pre-trained object detector to realize \textit{precise} localization of most video activities by detecting their associated subjects, e.g. humans. However, pre-trained detectors only detect objects in the \textit{``closed world''} formed by known object classes in the training dataset. This leads to a fatal problem: Anomalies, which are often novel classes outside the ``closed world'', will be omitted. To this end, motion cues like temporal gradients are proposed as complimentary information to accomplish more \textit{comprehensive} RoI localization. More importantly, we argue that appearance and motion cues should not be isolated: RoIs already localized by appearance should be filtered when exploiting motion cues, which reduces computation and makes motion based RoI localization more precise (see Fig. \ref{fig:visual_vee} (d)). As illustrated by the overview in Fig. \ref{fig:event_extraction} and Algorithm \ref{extract_roi}, we elaborate each component of the new video event extraction pipeline below.

\begin{algorithm}[t]
	\caption{Appearance and Motion based RoI Extraction}
	\label{extract_roi}
	\begin{algorithmic}[1] 
		\REQUIRE Frame $I$ and its gradient map $G$, pre-trained object detector $M$, threshold $T_s, T_a, T_o, T_g, T_{ar}$
		\ENSURE RoIs represented by a bounding box set $B$
		\STATE $B_{ap}\gets$ \textit{ObjDet}$(I, M, T_s)$ \quad \# Detecting activity subjects
		\STATE $B_{a}=\{\}$ \quad \# Heuristic filtering
		\FOR {$b_{ap} \in B_{ap}$}
		\mathchardef\mhyphen="2D
		\IF {$Area(b_{ap})>T_{a}$ \AND $Overlap(b_{ap}, B_{ap})<T_o$}
		\STATE $B_{a}=B_{a} \cup \{b_{ap}\}$
		\ENDIF
		\ENDFOR
		\STATE $G_b\gets$ \textit{GradBin}$(G, T_g)$ \quad \# Gradient binarization
		\STATE $G_b\gets$ \textit{RoISub}$(G_b, B_a)$ \quad \# Subtract appear. based RoIs
		
		\STATE $C\gets$ \textit{ContourDet}$(G_b)$ \quad \# Contour detection
		\STATE $B_{m}=\{\}$
		\FOR {$c \in C$}
		\STATE $b_{m}=BoundingBox(c)$ \quad \# Get contour bounding box
		\mathchardef\mhyphen="2D
		\IF {$Area(b_{m})>T_{a}$ \AND $\frac{1}{T_{ar}}<AspectRatio(b_{m})<T_{ar}$}
		\STATE $B_{m}=B_{m} \cup \{b_{m}\}$
		\ENDIF
		\ENDFOR
		\STATE $B=B_{a} \cup B_{m}$
	\end{algorithmic}
\end{algorithm}

\textbf{Appearance Based RoI Extraction.} Given a video frame $I$ and a pre-trained object detector model $M$, our goal is to obtain a RoI set $B_a$ based on appearance cues from subjects of video activities, where $B_a\subseteq \mathbb{R}^4$ and each entry of $B_a$ refers to a RoI enclosed by a bounding box. The bounding box is denoted by the coordinates of its top-left and bottom-right vertex, which is a 4-dimensional vector. As shown by the green module in Fig. \ref{fig:event_extraction}, we first feed $I$ into $M$, and obtain a preliminary RoI set $B_{ap}$ with confidence scores above the threshold $T_s$ (class labels are discarded). Then, we introduce two efficient heuristic rules to filter unreasonable RoIs: \textbf{(1)} RoI area threshold $T_a$ that filters out overly small RoIs. \textbf{(2)} Overlapping ratio $T_o$ that removes RoIs that are nested or significantly overlapped with larger RoIs in $B_{ap}$. In this way, we ensure that extracted RoIs can precisely enclose subjects of most everyday video activities.

\textbf{Motion Based RoI Extraction.} To enclose those activities outside the ``closed world'', motion based RoI extraction aims to yield a complementary bounding box set $B_m$ based on motion cues. We leverage the temporal gradients of frames as motion cues and complementary information. As shown by the red module of Fig. \ref{fig:event_extraction}, we first binarize the absolute values of temporal gradients by a threshold $T_g$, so as to yield a binary map that indicates regions with intense motion. Instead of using this map directly, we propose to subtract appearance based RoIs $B_a$ from the map, which benefits motion based RoI extraction in two ways: First, the subtraction of appearance based RoIs enables us to better localize objects that are not detected by appearance cues, otherwise the gradient map of multiple objects may be overlapped and jointly produce large and imprecise RoIs (see Fig. \ref{fig:visual_vee} (b)). Second, the subtraction reduces the computation. Finally, we propose to perform contour detection to yield the contour and its corresponding bounding box $b_m$, while simple heuristics (RoI area threshold $T_a$ and maximum aspect-ratio threshold $T_{ar}$) are used to obtain final RoI set $B_m$. Based on two complementary RoI sets, the final RoI set $B=B_a\cup B_m$. The whole RoI extraction process is formally presented in Algorithm \ref{extract_roi}.

\textbf{Spatio-temporal Cube Construction.} Finally, we use each RoI in $B$ to build a spatio-temporal cube (STC) as the video event, which represents the fundamental unit to enclose video activities. As shown by yellow module in Fig. \ref{fig:event_extraction}, we not only extract the patch $p_1$ in the RoI from current frame, but also extract corresponding patches $p_2,\cdots, p_D$ by this RoI from previous $(D-1)$ frames. In this way, we incorporate the temporal context into the extracted video event. To normalize video activities with different scales, we resize patches from the same RoI into $H\times W$ new patches $p'_1, \cdots, p'_D$, which are then stacked into a $H\times W\times D$ STC: $C=[p'_1; \cdots; p'_D]$.

\subsection{Visual Cloze Tests}
As explained by \textbf{Gap \#2} in Sec. \ref{sec:intro}, previous methods typically rely on a reconstruction or frame prediction paradigm. They cannot fully exploit high-level semantics and temporal context information. As a remedy, we propose to build \textit{visual cloze tests} as a new learning paradigm, so as to model normal video activities represented by STCs above. We present it in terms of the following aspects:
\begin{figure*}
	\centering
	\includegraphics[scale=1.08]{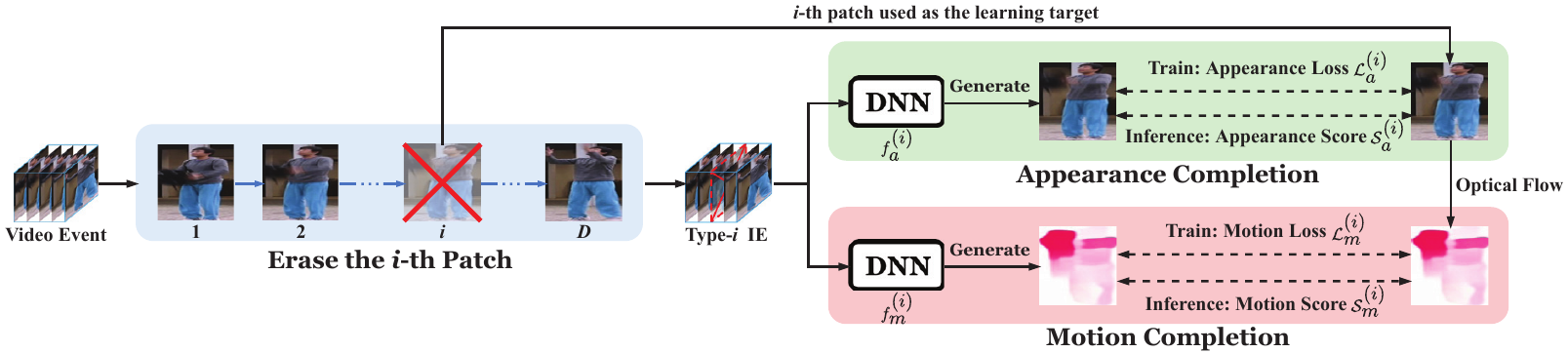}
	\caption{Visual cloze tests with a type-$i$ incomplete event (IE): \textbf{(1)} Erase the $i$-th patch (blue): The $i$-th patch of a STC is erased to build a type-$i$ IE, while the erased patch is used as the learning target of appearance completion. \textbf{(2)} Appearance completion (green): To complete the IE, a DNN takes the IE as input and learns to generate the erased patch. \textbf{(3)} Motion completion (red): A DNN takes the IE as input and learns to generate the optical flow patch that corresponds to the erased patch.}
	\label{fig:event_completion}
\end{figure*}

\textbf{Motivation.}
We are inspired by \textit{cloze test}, an extensively-used exercise in language learning and instruction. It requires completing a text with its certain words erased. Cloze test aims to test students' ability to understand the vocabulary semantics and language context \cite{wiki_cloze_test}. Recently, learning to solve cloze tests is also shown to be a fairly effective pretraining method in natural language processing (NLP), which enables DNN to capture richer high-level semantics from text \cite{devlin2019bert}. This naturally inspires us to compare the patch sequence of a STC to the word sequence in classic cloze test. Similarly, we can erase a certain patch $p'_i$ in video event (STC) to build a visual cloze test, which is solved by completing the resultant \textit{incomplete event} (IE) with the DNN's inferred $\Tilde{p}'_i$. Such a learning paradigm benefits DNN in two aspects: \textbf{(1)} In order to complete such a visual cloze test, DNN is encouraged to capture high-level semantics in STC. For example, suppose a video event contains a walking person. DNN must attend to those key moving parts (e.g. the forwarding leg and swinging arm) in the patch to achieve a good completion. This makes visual cloze test more meaningful than frequently-used reconstruction, which tends to memorize every low-level detail of inputs to minimize training loss. \textbf{(2)} Since any patch in a STC can be erased to generate an IE, we can build multiple cloze tests by erasing the patch at different temporal positions. This enables us to fully exploit the temporal context by enumerating all possible IEs for cloze tests. By contrast, prediction based VAD methods only consider prediction errors of a single frame to detect anomalies, which involves poor temporal context information in video activities. As shown in Fig. \ref{fig:event_completion}, we detail VEC's components below.

\textbf{Appearance Completion.}
Given $j$-th video event represented by the STC $C_j=[p'_{j,1}; \cdots; p'_{j,D}]$, we first erase the $i$-th patch $p'_{j,i}$ of $C_j$ to build an IE $C_j^{(i)}=[p'_{j,1};\cdots p'_{j,i-1}; p'_{j,i+1};\cdots p'_{j,D}]$ as a cloze test, $i\in \{1,\cdots D\}$ (blue module in Fig. \ref{fig:event_completion}). All IEs built by erasing the $i$-th patch of the STC are collected as the type-$i$ IE set $\mathcal{C}^{(i)}=\{C_1^{(i)}, \cdots C_N^{(i)}\}$, where $N$ is the number of extracted video events (STCs). Afterwards, as shown by red module in Fig. \ref{fig:event_completion}, a type-$i$ IE $C_j^{(i)}$ in $\mathcal{C}^{(i)}$ and its corresponding erased patch ${p}'_{j,i}$ are used as the input and learning target respectively to train a generative DNN $f^{(i)}_a$ (e.g. autoencoder, generative adversarial networks, U-Net, etc.), which aims to generate a patch $\Tilde{p}'_{j,i}=f^{(i)}_a(C_j^{(i)})$ to complete the IE ${C}_j^{(i)}$ into the original event $C_j$. To train $f^{(i)}_a$, here we can simply minimize the pixel-wise appearance loss $\mathcal{L}^{(i)}_a$ for type-$i$ IE set $\mathcal{C}^{(i)}$:
\begin{equation}
\mathcal{L}^{(i)}_a = \frac{1}{N}\sum^N_{j=1}\Vert \Tilde{p}'_{j,i}-{p}'_{j,i}\Vert^p_p
\end{equation}
where $\Vert\cdot\Vert_p$ denotes $p$-norm. In our experiments, we found that choosing $p=2$ already works well. To further improve the fidelity of generated patch, other values of $p$ or techniques like adversarial training can also be explored to design $\mathcal{L}^{(i)}_a$. For inference, any error measure $\mathcal{S}^{(i)}_a(\Tilde{p}'_{j,i}, {p}'_{j,i})$ can be used to yield the appearance anomaly score of patch ${p}'_{j,i}$, such as mean square error (MSE) or Peak Signal to Noise Ratio (PSNR) \cite{liu2018future}. Empirically, choosing $\mathcal{S}^{(i)}_a(\Tilde{p}'_{j,i}, {p}'_{j,i})$ to be simple MSE has been effective enough to score anomalies, and poorly completed STCs with higher MSE are more likely to be anomalies. Since appearance completion is actually a challenging learning task for DNN, an independent DNN $f^{(i)}_a$ is trained to handle one IE type $\mathcal{C}^{(i)}$. Otherwise, using one DNN for all IE types will degrade the performance of appearance completion.

\textbf{Motion Completion.}
Motion is another type of valuable prior in videos. Optical flow, which estimates the pixel-wise motion velocity and direction between two consecutive frames, is a popular low-level motion representation in videos. We feed two consecutive frames into a pre-trained FlowNet model \cite{ilg2017flownet}, and a forward pass can yield the optical flow efficiently. For each STC $C_j$, we extract optical flow patches $o_{j,1}, \cdots o_{j,D}$ that correspond to video patches $p_{j,1}, \cdots p_{j,D}$, and also resize them into $H\times W$ patches $o'_{j,1}, \cdots o'_{j,D}$. Motion completion requires a DNN $f^{(i)}_m$ to infer the optical flow patch $o'_{j,i}$ of the erased patch $p'_{j,i}$ by the type-$i$ IE $C_j^{(i)}$, i.e. $\Tilde{o}'_{j,i}=f^{(i)}_m(C_j^{(i)})$. $f^{(i)}_m$ can be trained by the motion loss $\mathcal{L}^{(i)}_m$:

\begin{equation}
\mathcal{L}^{(i)}_m = \frac{1}{N}\sum^N_{j=1}\Vert \Tilde{o}'_{j,i}-{o}'_{j,i}\Vert^p_p
\end{equation}
Likewise, we also adopt $p=2$ for $\mathcal{L}^{(i)}_m$ and simple MSE to compute the motion anomaly score $\mathcal{S}^{(i)}_m(\Tilde{o}'_{j,i}, {o}'_{j,i})$ during inference. With motion completion, we encourage DNN to infer the motion statistics from the temporal context provided by IEs, which enables VEC to consider richer motion dynamics. The process of both appearance and motion completion for a type-$i$ IE are shown in Fig. \ref{fig:event_completion}.

\textbf{Ensemble Strategies.}
Ensemble is a powerful tool that combines multiple models into a stronger one \cite{dietterich2000ensemble}. During inference, we propose two ensemble strategies to improve the VEC performance: \textbf{(1)} \textit{IE type ensemble}. Erasing a different patch in STC produces a different IE, which contains a different patch combination as temporal context. To fully exploit all possible temporal context for VAD, we compute the final appearance anomaly score by an ensemble of scores, which are yielded by multiple DNNs for different IE types:

\begin{equation}
\mathcal{S}_a(C_j) = \frac{1}{D}\sum^D_{i=1}\mathcal{S}^{(i)}_a(\Tilde{p}'_{j,i}, {p}'_{j,i})
\end{equation}
IE type ensemble is also applicable to the final motion score $\mathcal{S}_m(C_j)$:
\begin{equation}
\mathcal{S}_m(C_j) = \frac{1}{D}\sum^D_{i=1}\mathcal{S}^{(i)}_m(\Tilde{o}'_{j,i}, {o}'_{j,i})
\end{equation}
\textbf{(2)} \textit{Modality ensemble}. Since two different modalities, raw pixels and optical flow, are considered to perform completion in VEC, we must fuse their results to yield the overall anomaly score. For simplicity, we use a weighted sum of $\mathcal{S}_a(C_j)$ and $\mathcal{S}_m(C_j)$ to compute the overall anomaly score $S(C_j)$ for a video event $C_j$:

\begin{equation}
\mathcal{S}(C_j)=w_a\cdot\frac{\mathcal{S}_a(C_j)-\Bar{\mathcal{S}}_a}{\sigma_a} + w_m\cdot\frac{\mathcal{S}_m(C_j)-\Bar{\mathcal{S}}_m}{\sigma_m}
\end{equation}
where $\Bar{\mathcal{S}}_a, \sigma_a, \Bar{\mathcal{S}}_m, \sigma_m$ denote the means and standard deviations of appearance and motion scores for all normal events in training, which are used to normalize appearance and motion scores into the same scale. In addition to this straightforward weighting strategy, other more sophisticated strategies like late fusion \cite{xu2017detecting} are also applicable to achieve better modality ensemble performance.

\section{Evaluation}
\label{sec:eval}

\subsection{Experimental Settings}
\begin{figure}
	\centering
	\includegraphics[scale=1.12]{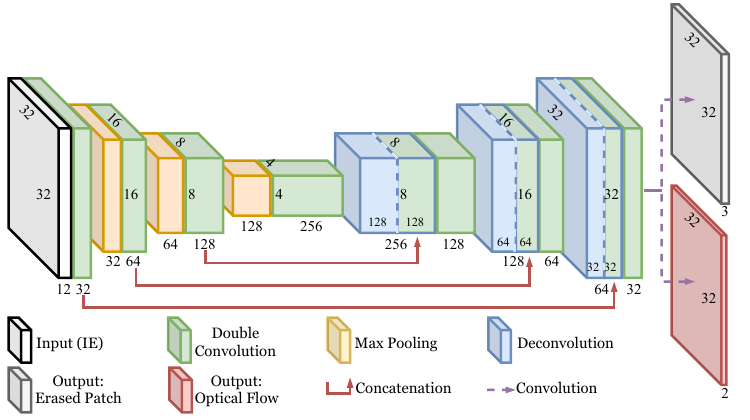}
	\caption{DNN architecture used in our experiments. Appearance completion network $f^{(i)}_a$ and motion completion network $f^{(i)}_m$ share the same U-Net architecture, except that $f^{(i)}_a$ has 3 output channels (images) while $f^{(i)}_m$ has 2 (optical flow).}
	\label{fig:U-Net}
\end{figure}

Evaluation is performed on three most commonly-used benchmark datasets for VAD: UCSDped2 \cite{mahadevan2010anomaly}, Avenue \cite{lu2013abnormal} and ShanghaiTech \cite{luo2017revisit}. For video event extraction, cascade R-CNN \cite{cai2018cascade} pre-trained on COCO dataset is used as object detector as it achieves a good trade-off between performance and speed. Other parameters are set as follows for UCSDped2, Avenue and ShanghaiTech respectively: Confidence score threshold $T_s$: $(0.5, 0.25, 0.5)$; RoI area threshold $T_a$: $(10\times10, 40\times40, 8\times8)$; Overlapping ratio $T_o$: $(0.6, 0.6, 0.65)$; Gradient binarization threshold $T_g$: $(18, 18, 15)$; Maximum aspect-ratio threshold $T_{ar}=10$. For cube construction, we set $H=W=32$ and $D=5$. As to VEC, we adopt U-Net \cite{ronneberger2015u} as the basic network architecture of generative DNNs (see Fig. \ref{fig:U-Net}), which are optimized by the default Adam optimizer in PyTorch \cite{paszke2019pytorch}. Considering the dataset scale, DNNs are trained by 5, 20, 30 epochs with a batch size 128 on UCSDped2, Avenue and ShanghaiTech respectively. For anomaly scoring, we set $(w_a, w_m)$ to be $(0.5, 1)$, $(1, 1)$ and $(1, 0.5)$ for UCSDped2, Avenue and ShanghaiTech respectively. For quantitative evaluation, we adopt the most frequently-used metric: Area Under the Receiver Operating Characteristic curves (AUROC) that are computed with frame-level detection criteria \cite{mahadevan2010anomaly}. Frame-level Equal Error Rate (EER) \cite{mahadevan2010anomaly} is also reported in the supplementary material. We run experiments on a PC with 64 GiB RAM, Nvidia Titan Xp GPUs and a 3.6GHz Intel i7-9700k CPU.

\subsection{Comparison with State-of-the-Art methods}

\begin{figure*}[t]
	\centering
	\includegraphics[scale=0.68]{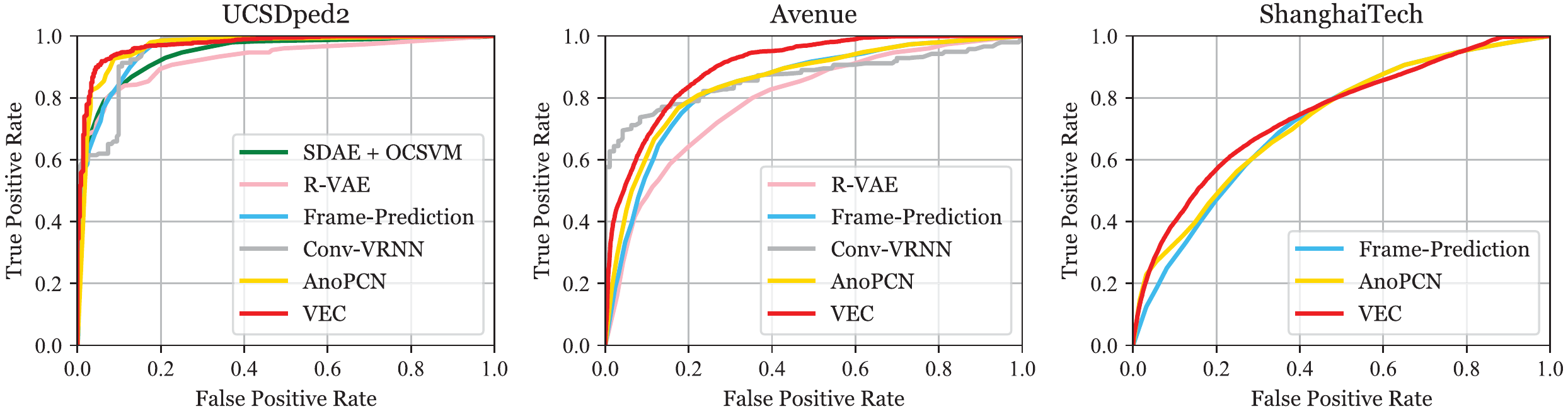}
	\caption{Comparison of frame-level ROC curves on different VAD benchmark datasets.}
	\label{fig:ROC}
\end{figure*}

Within our best knowledge, we extensively compare VEC's performance with 18 state-of-the-art DNN based VAD methods reviewed in Sec. \ref{sec:related_work}. Note that we exclude \cite{ionescu2019object} from comparison as it actually uses a different evaluation metric from commonly-used frame-level AUROC, which leads to an unfair comparison. As discussed in Sec. \ref{sec:related_work}, existing methods can be categorized into reconstruction based, frame prediction based and hybrid methods in Table \ref{tab:comp_stoa}. As to VEC, we design two configurations for IE type ensemble: \textbf{(1)} VEC-A: IE type ensemble is applied to appearance completion, while it is not applied to motion completion (i.e. only type-$D$ IEs are used to train the DNN for motion completion). \textbf{(2)} VEC-AM: IE type ensemble is applied to both appearance completion and motion completion. Besides, modality ensemble is applied to both VEC-A and VEC-AM. The results are reported in Table \ref{tab:comp_stoa}, and we visualize the yielded frame-level ROC curves in Fig. \ref{fig:ROC}. We draw the following observations: First, both VEC-A and VEC-AM consistently outperform existing state-of-the-art DNN based VAD methods on three benchmarks. In particular, we note that VEC achieves notable performance gain on recent challenging benchmarks (Avenue and ShanghaiTech) with constantly $\geq 3\%$ and $\geq 1\%$ AUROC improvement respectively against all state-of-the-art methods. Meanwhile, we note that VEC-A even achieves over $90\%$ frame-level AUROC on Avenue, which is the best performance ever achieved on Avenue dataset to our knowledge. In terms of the comparison between VEC-A and VEC-AM, two configurations yield fairly close performance, despite of slight differences on different datasets. As a consequence, in addition to those thoroughly-studied reconstruction or prediction based methods, the proposed VEC provides a highly promising alternative for DNN based VAD with state-of-the-art performance.

\begin{table*}
	\newcommand{\tabincell}[2]{\begin{tabular}{@{}#1@{}}#2\end{tabular}}
	\centering
	\linespread{1}\selectfont
	\scalebox{1}{
		\begin{tabular}{ccccc}
			\toprule
			\multicolumn{2}{c}{Method}                   & UCSDped2  & Avenue  & ShanghaiTech\\
			\midrule
			\multirow{11}{*}{\tabincell{c}{Reconstruction\\Based Methods}} & CAE \cite{hasan2016learning}  & 85.0\%  & 80.0\%  & 60.9\%   \\
			&SDAE + OCSVM \cite{xu2017detecting}                  & 90.8\% &  -  & -   \\
			&SRNN \cite{luo2017revisit}                 & 92.2\%  & 81.7\%  & 68.0\%   \\
			&GAN \cite{ravanbakhsh2017abnormal}                 & 93.5\%  & -  & -   \\
			&ConvLSTM-AE \cite{luo2017remembering}      & 88.1\% & 77.0\% & - \\
			&WTA-CAE + OCSVM \cite{tran2017anomaly}          & 96.6\% & 82.1\% & - \\
			&R-VAE \cite{yan2018abnormal}           & 92.4\% &79.6\% & - \\
			&PDE-AE \cite{abati2019latent}               & 95.4\% & - & 72.5\% \\
			&Mem-AE \cite{Gong_2019_ICCV}      & 94.1\%   & 83.3\%  & 71.2\%     \\
			&AM-CORR \cite{nguyen2019anomaly}    & 96.2\% & 86.9\%     & - \\
			&AnomalyNet \cite{zhou2019anomalynet}  & 94.9\% & 86.1\% & - \\
			\hline
			\multirow{3}{*}{\tabincell{c}{Frame Prediction\\Based Methods}}
			&Frame-Prediction \cite{liu2018future}      & 95.4\%   & 84.9\%  & 72.8\%     \\
			&Conv-VRNN \cite{lu2019future}    & 96.1\% & 85.8\% & - \\
			&Attention-Prediction \cite{zhou2019attention}  & 96.0\% & 86.0\% & - \\
			\hline
			\multirow{4}{*}{\tabincell{c}{Hybrid Methods}}
			&ST-CAE \cite{zhao2017spatio}                  & 91.2\% & 80.9\%  & -   \\
			&Skeleton-Trajectories + MPED-RNN \cite{morais2019learning}    & - & - & 73.4\% \\
			&AnoPCN \cite{ye2019anopcn}      & 96.8\%   & 86.2\%  & 73.6\%     \\
			&Prediction$\&$Reconstruction \cite{tang2020integrating} & 96.3\%  & 85.1\%  & 73.0\%   \\
			\hline
			\multirow{2}{*}{\textbf{Proposed}}
			& VEC-A      & 96.9\% &\textbf{90.2\%} & 74.7\% \\
			& VEC-AM     &\textbf{97.3\%} &89.6\% &\textbf{74.8\%}\\
			\bottomrule
	\end{tabular}}
	\caption{AUROC comparison between the proposed VEC and state-of-the-art VAD methods.}
	\label{tab:comp_stoa}
\end{table*}

\subsection{Detailed Analysis}

\begin{table}
	\centering
	\linespread{1}\selectfont
	\scalebox{0.89}{
		\begin{tabular}{cccccccc}
			\toprule
			\multirow{2}{*}{Dataset} & \multicolumn{4}{c}{Video Event Extraction} & \multicolumn{2}{c}{Ensemble} & \multirow{2}{*}{AUROC}\\
			& FR & SDW & APR & APR+MT & IE Type & Modality & \\
			\midrule
			\multirow{6}{*}{\rotatebox{90}{UCSDped2}} & \CheckmarkBold & & & & \CheckmarkBold & \CheckmarkBold & 94.6\% \\
			& & \CheckmarkBold & & & \CheckmarkBold &\CheckmarkBold & 93.3\% \\
			& & & \CheckmarkBold & & \CheckmarkBold &\CheckmarkBold & 95.5\% \\
			& & & & \CheckmarkBold &  & \CheckmarkBold & 96.0\% \\
			& & & & \CheckmarkBold & \CheckmarkBold &  & 89.6\% \\
			& & & & \CheckmarkBold & \CheckmarkBold & \CheckmarkBold & \textbf{97.3\%} \\
			\hline
			\multirow{6}{*}{\rotatebox{90}{Avenue}} & \CheckmarkBold & & & & \CheckmarkBold & \CheckmarkBold & 86.8\% \\
			& & \CheckmarkBold & & & \CheckmarkBold & \CheckmarkBold & 85.2\% \\
			& & & \CheckmarkBold & & \CheckmarkBold & \CheckmarkBold & 87.1\% \\
			& & & & \CheckmarkBold &  & \CheckmarkBold & 87.5\% \\
			& & & & \CheckmarkBold & \CheckmarkBold &  & 88.2\% \\
			& & & & \CheckmarkBold & \CheckmarkBold & \CheckmarkBold & \textbf{89.6\%} \\
			\hline
			\multirow{6}{*}{\rotatebox{90}{ShanghaiTech}} & \CheckmarkBold & & & & \CheckmarkBold & \CheckmarkBold & 70.2\% \\
			& & \CheckmarkBold & & & \CheckmarkBold & \CheckmarkBold & - \\
			& & & \CheckmarkBold & & \CheckmarkBold & \CheckmarkBold &  73.6\%\\
			& & & & \CheckmarkBold &  & \CheckmarkBold & 74.4\% \\
			& & & & \CheckmarkBold & \CheckmarkBold &  &  73.5\%\\
			& & & & \CheckmarkBold & \CheckmarkBold & \CheckmarkBold & \textbf{74.8\%} \\
			\bottomrule
	\end{tabular}}
	\caption{Ablation Studies for VEC.}
	\label{tab:ablation}
\end{table}

\textbf{Ablation Studies.} To show the role of the proposed video event extraction and ensemble strategies, we perform corresponding ablation studies and display the results in Table \ref{tab:ablation}: \textbf{(1)} As to video event extraction, we compare four practices for localizing video activities: Frame (FR, i.e. no localization at all), multi-scale sliding windows with motion filtering (SDW), appearance based RoI extraction only (APR) and the proposed appearance and motion based RoI extraction (APR+MT). Note that we did not report SDW's results on ShanghaiTech, since it produces excessive STCs that are beyond the limit of our hardware, which is actually an important downside of SDW. There are several observations: First, with the same ensemble strategies, the proposed APR+MT constantly outperforms other methods by a sensible margin. Specifically, APR+MT has an obvious advantage ($2.7\%$-$4.6\%$ AUROC gain) over FR and SDW, which are commonly-used strategies of recent VAD methods. Interestingly, we note that SDW performs worse than FR on UCSDped2 and Avenue. This indicates that an imprecise localization of video activities even degrades VAD performance, and it also justifies the importance of a precise localization. Meanwhile, when compared with APR, the proposed APR+MT brings evident improvement by $1.8\%$, $2.5\%$ and $1.2\%$ AUROC gain on UCSDped2, Avenue and ShanghaiTech respectively. Such observations demonstrate that a more comprehensive localization of video activities will contribute to VAD performance. \textbf{(2)} As to ensemble strategies, we compare three cases: Not using IE type ensemble (for both appearance and motion completion), not using modality ensemble ($w_m=0$), and both IE type and modality ensemble are used (VEC-AM). We yield the following observations: First, IE type ensemble contributes to VAD performance by $1.3\%$, $2.1\%$ and $0.4\%$ AUROC on UCSDped2, Avenue and ShanghaiTech respectively, which justifies the importance to fully exploit temporal context. Second, modality ensemble enables a remarkable $8\%$ AUROC gain on UCSDped2. This is because UCSDped2 contains low-resolution gray-scale frames, and motion clues are more important for detecting anomalies. For Avenue and ShanghaiTech with high-resolution colored frames, modality ensemble also enables over $1\%$ AUROC improvement, although using the modality of raw pixel only already leads to satisfactory performance.

\begin{figure}
	\centering
	\includegraphics[scale=0.75]{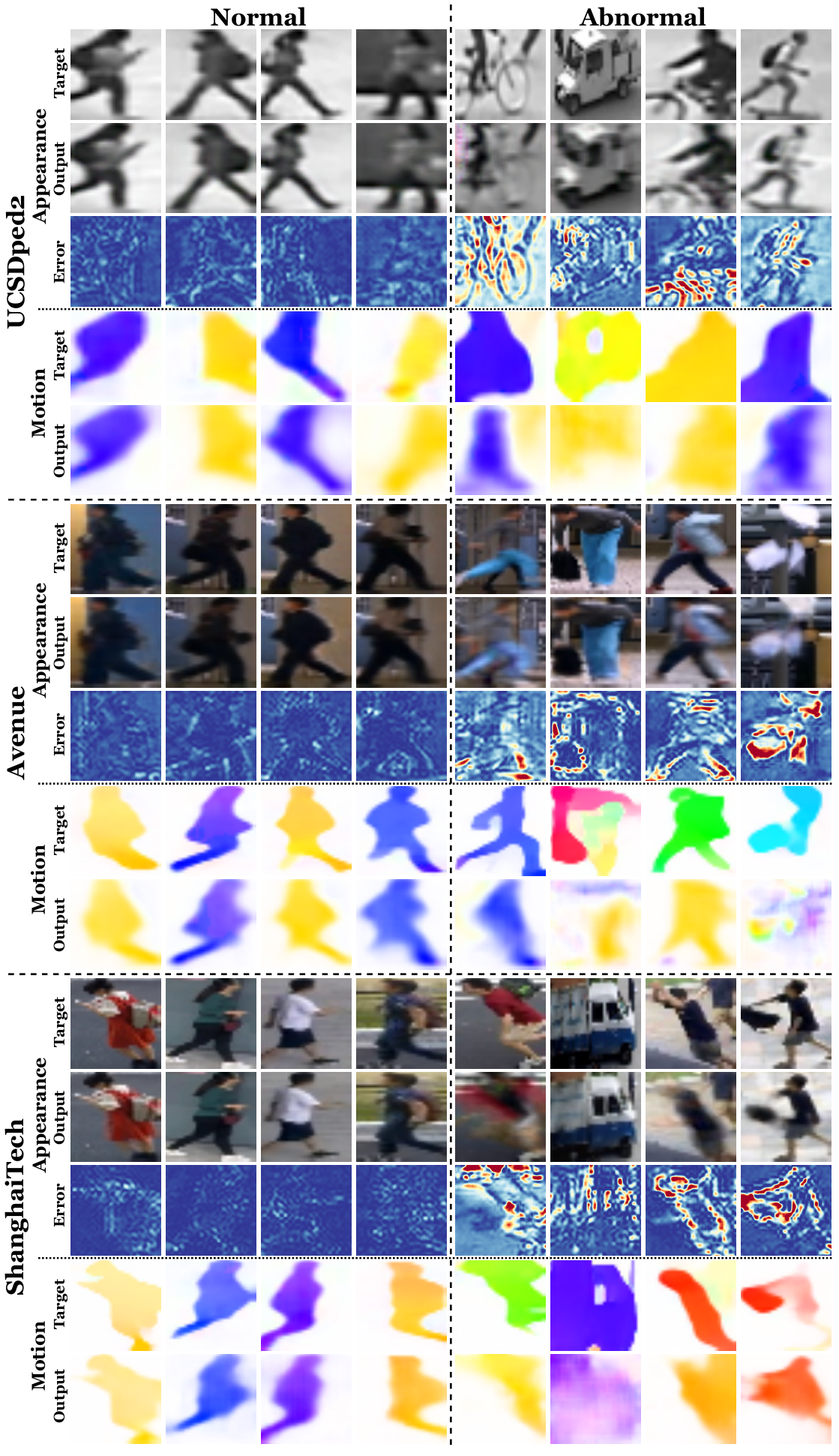}
	\caption{Visualization of erased patches and their optical flow (Target), completed patches (Output) by VEC and completion errors (Error). Brighter color indicates larger errors.}
	\label{fig:visual}
\end{figure}

\textbf{Visualization.} To show how visual cloze tests in VEC helps discriminating anomalies in a more intuitive way, we visualize generated patches and optical flow of representative normal/abnormal video events in Fig. \ref{fig:visual}. Heat maps are used for a better visualization of the pixel-wise completion errors. By Fig. \ref{fig:visual}, it is worth noting several phenomena: First of all, VEC can effectively complete normal events and their optical flow. For normal events, minor completion errors are observed to be distributed around foreground contour in a relatively uniform manner, and their optical flow can also be soundly recovered. By contrast, abnormal events produces prominent completion errors in terms of both raw pixel and optical flow completion. Next, it is noted that the distribution of anomalies' completion errors is highly non-uniform. As shown by heat maps, large completion errors are often observed at those regions that have clear high-level semantics, e.g. the bicycle that the man was riding with (UCSDped2), falling paper with its shape and position wrongly inferred (Avenue), the backpack that was thrown (ShanghaiTech). By contrast, other regions are endowed with relatively smaller errors. Such observations imply that VEC indeed attends to those parts with high-level semantics in abnormal events. 

\textbf{Other Remarks.} \textbf{(1)} VEC adopts a similar U-Net architecture to previous works, but it achieves significantly better performance, which exactly verifies visual cloze tests' effectiveness as a new learning paradigm. Thus, better network architecture can be explored, while techniques like adversarial training and attention mechanism are also applicable. In fact, VEC with only type-$D$ IEs can be viewed as predicting the last patch of STCs (as shown in Table \ref{tab:ablation}, it is also better than frame prediction \cite{liu2018future}). Besides, when visual cloze tests in VEC are replaced by plain reconstruction, experiments report a $3\%$ to $7\%$ AUROC loss on benchmarks, which demonstrates that our video event extraction and visual cloze tests are both indispensable. \textbf{(2)} Details of VEC's computation cost and parameter sensitivity are also discussed in the supplementary material.

\section{Conclusion}
\label{sec:conclusion}
In this paper, we propose VEC as a new solution to DNN based VAD. VEC first extracts STCs by exploiting both appearance and motion cues, which enables both precise and comprehensive video event extraction. Subsequently, motivated by the widely-used cloze test, VEC learns to solve visual cloze tests, i.e. training DNNs to infer deliberately erased patches from incomplete video events/STCs, so as to learn better high-level semantics. Motion modality is also involved by using DNNs to infer the erased patches' optical flow. Two ensemble strategies are further adopted to fully exploit temporal context and motion dynamics, so as to enhance VAD performance.

\begin{acks}	
The work is supported by National Natural Science Foundation of China under Grant No. 61702539, Hunan Provincial Natural Science Foundation of China under Grant No. 2018JJ3611, No. 2020JJ5673, NUDT Research Project under Grant No. ZK-18-03-47, ZK20-10, and The National Key Research and Development Program of China (2018YFB0204301, 2018YFB1800202, SQ2019ZD090149). Siqi Wang, Zhiping Cai and Jianping Yin are corresponding authors.
\end{acks}
\bibliographystyle{ACM-Reference-Format}
\bibliography{mmfp0444}

\end{document}